\pdfoutput=1
\documentclass[11pt]{article}

\usepackage[]{acl}
\usepackage{amsmath}
\usepackage[ruled]{algorithm2e}
\usepackage{times}
\usepackage{latexsym}
\usepackage{svg}
\usepackage[T1]{fontenc}
\usepackage[utf8]{inputenc}

\usepackage{microtype}

\usepackage{inconsolata}
\usepackage{ragged2e}
\usepackage{mdframed}
\usepackage{graphicx}
\usepackage{subcaption}
\usepackage{longtable}
\usepackage{multirow}
\usepackage[table]{colortbl}% http://ctan.org/pkg/xcolor
\usepackage{tikz}
\usepackage{caption}
\usepackage{booktabs}
\usetikzlibrary{positioning}
\usetikzlibrary{arrows.meta}

\newcommand{\factscore}{\textsc{FActScore}}
\newcommand{\decompscore}{\textsc{DecompScore}}
\newcommand{\veriscore}{\textsc{VeriScore}}

\newcommand{\wice}{\textsc{WiCE}}
\newcommand{\faithscore}{\textsc{FaithScore}}
\newcommand{\decompmethod}{$\mathcal{D}_{\scriptsize \textnormal{R-ND}}$}
\newcommand{\newscore}{\textsc{DnDScore}}
\newcommand{\decompdecontext}{DnD}
\newcommand{\core}{\textsc{Core}}

\newcolumntype{M}[1]{>{\centering\arraybackslash}m{#1}}

\title{\includegraphics[scale=0.055]{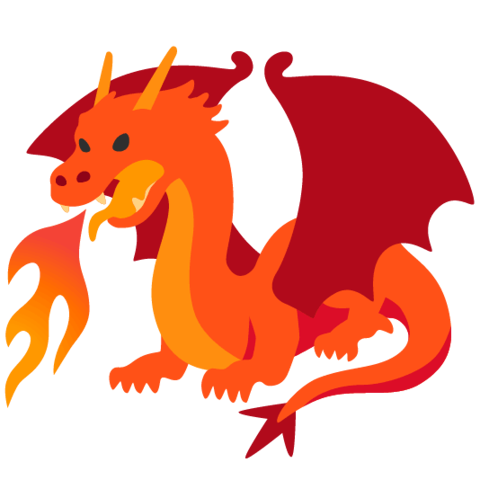} \newscore{}: Decontextualization and Decomposition for \\Factuality Verification in Long-Form Text Generation}
\author{Miriam Wanner, Benjamin Van Durme, Mark Dredze\\[1em] Johns Hopkins University \\[1em] 
\texttt{\{mwanner5,vandurme,mdredze\}@jhu.edu} \\ [1em] }
\date{}

\begin{document}

\maketitle

\begin{abstract}
The decompose-then-verify strategy for verification of Large Language Model (LLM) generations decomposes claims that are then independently verified.
Decontextualization augments text (claims) to ensure it can be verified outside of the original context, enabling reliable verification. While
decomposition and decontextualization have been explored independently, 
their interactions in a complete system have not been investigated. Their conflicting purposes can create tensions: decomposition isolates atomic facts while decontextualization inserts relevant information. Furthermore, a 
decontextualized subclaim presents a challenge to the verification step: what part of the augmented text should be verified as it now contains
multiple atomic facts?
We conduct an evaluation of different decomposition, decontextualization, and verification strategies and find that the choice of strategy matters in the resulting factuality scores. Additionally, we introduce \newscore{}, a decontextualization aware verification method which validates subclaims in the context of contextual information.
\end{abstract}

\section{Introduction}
\begin{figure*}
    \centering
    \includegraphics[width=1\linewidth]{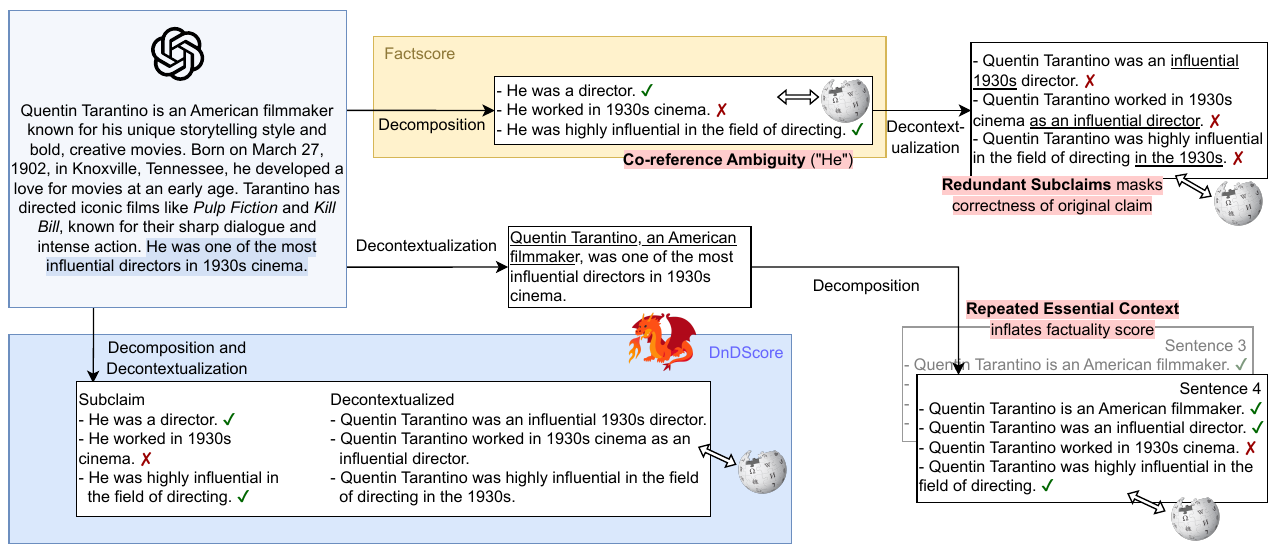}
    \caption{Current claim verification methods evaluate subclaims out of context, but adding this context into the claim verification pipeline is not trivial. We introduce \newscore{}, a method that evaluates subclaims given the decontextualized claim or another context to help verify.}
    \label{fig:motivation}
\end{figure*}

Factuality evaluations measure the correctness of language model generations. Recent measures of factual precision utilize a decompose-then-verify framework, where text is first decomposed into atomic subclaims and then validated against a trusted source document \citep{min2023factscore, jiang2024corerobustfactualprecision}. However, decomposition may remove information necessary to understand the claim. For example, in Figure \ref{fig:motivation}, the decomposition of the sentence ``He was one of the most influential directors in 1930s cinema.'' would include the subclaim ``He was a director.'', which lacks entity context (who does ``He'' refers to?). These ambiguities prevent successful claim verification, and could lead to false positives.

Decontextualization is the process of augmenting subclaims with necessary context to an ensure an accurate claim verification independent of the rest of the generation. Decontextualization can include pronoun replacement, name completion, or addition of information from the original text \citep{choi-etal-2021-decontextualization}. Decontextualizing decomposed atomic subclaims provides necessary information, but introduces several other problems. By introducing new information to the subclaim, the claim becomes less atomic, making it unclear which part of the new claim requires verification. For example, in Figure \ref{fig:motivation}, decontextualizing ``He was a director.'' results in ``Quentin Tarantino was an influential 1930s director''. As a result, the verification task changes from validating whether Tarantino was a directer, to validating whether he was a \textit{1930s} director. In addition, the potential incorrectness of information added at the decontextualization step can mask the correctness of the original subclaim. Finally, by decontextualizing a set of decomposed claims, we are at risk of building a set of redundant claims that closely resemble the original claim (as shown in top right set of claims in Figure \ref{fig:motivation}). Decontextualization alone still may not solve the original problem, where the subclaim does not stand alone from the original context and therefore cannot be verified.

Issues with factuality evaluation persist even if we flip the verification pipeline from decompose-then-decontextualize to decontextualize-then-decompose, where the original claim is first decontextualized, then decomposed, and finally verified. This approach may result in verifying repeated context across claims, inflating the resulting factuality score. For example in the bottom right of Figure \ref{fig:motivation}, the subclaim ``Quentin Tarantino is an American filmmaker.'' is generated for both Sentence 3 and Sentence 4. Additionally, decomposing the decontextualized claim results in a loss of context for every subclaim added in the decontextualization step. If there was another person named Quentin Tarantino, we would need additional information to disambiguate which Quentin Tarantino was being discussed. Current research has validated the decomposition \citep{wanner-etal-2024-closer} and decontextualization \citep{gunjal2024desiderata} processes independently, but have not considered these interactions.

We evaluate three methods for performing both decomposition and decontextualization in the same system. First, we consider decomposition then decontextualization, which enables us to first isolate the atomic subclaims and then add the relevant context. We next consider decontextualization then decomposition, where we ensure claims have necessary context before decomposing them into subclaims. Finally, we propose a new joint method, \decompdecontext{}, which jointly performs decomposition and deontextualization.

Since the resulting decontextualized claim is no longer atomic, it presents an inherent ambiguity to the verified: which portion of the claim is being validated?
We propose a new verification method \newscore{} (Decomposition and Decontextualization Score) that considers both an atomic subclaim and its decontextualized form for verification, indicating the specific claim to verify and its relevant context. 
We find that using our new contextualized fact verification method changes if claims are verified, highlighting problems with current decompose-then-verify scores. We also show examples and common settings where \newscore{} and decompose-then-verify methods differ, and where claims are easier or more difficult to verify.

\section{Related Works}

\subsection{Decomposition}
There are several systems that implement the decompose-then-verify framework, varying in decomposition and verification strategy. \citet{min2023factscore} introduce \factscore{}, which utilizes a LLM-prompted decomposition step for claim verification. \citet{jing2023faithscore} use a similar pipeline, but introduce  \faithscore{}, a variation for evaluating the output of vision-language models. \citet{song-etal-2024-veriscore} introduce a new flavor of \factscore{}, called \veriscore{}, which extracts and verifies only claims which are verifiable. They include a contextualized method with a sliding window for verifiable claim extraction, but do not use context in the verification step.

Several studies have tried alternate strategies to factuality scoring. \citet{kamoi-etal-2023-wice} introduce a new textual entailment dataset, \wice{}, and use an automatic LLM Claim-Split approach, separating claims into sub-sentence units used for entailment classification. Using both the dataset and decomposition strategy, they tackle verification and retrieval. \citet{chen-etal-2024-complex} generate yes/no questions aligning to specific aspects of a claim for claim verification. A similar system is built in \citet{chen-etal-2022-generating}, also incorporating implied sub-questions. \citet{chen2023sub} introduce a sub-sentence encoder, a contextual embedding model that creates distinct embeddings for atomic facts within a sentence, and show its use in supporting fact retrieval and text attribution. \citet{li-etal-2024-self} introduce \textsc{Self-Checker}, a factuality verification method which extracts claims, generates queries from those claims, which are then used for retrieval and ultimately verification of claims. \citet{tang-etal-2024-minicheck} propose an efficient method for fact-checkin LLMs called MiniCheck. They construct GPT-4 synthetic training data to train a sentence-level fact-checker.

We follow the decompose-then-verify framework, where the decomposition step can have a significant effect on downstream factuality scores. 
We build on previous work that explores different decomposition methods. We use the decomposition technique of 
\citet{wanner-etal-2024-closer}, who propose a method grounded in Russellian and Neo-Davidsonian theory, and show its benefits over other methods.
\core{} is a method for subclaim selection that filters based on uniqueness and informativeness \citep{jiang2024corerobustfactualprecision}. \citet{hu2024decompositiondilemmasdoesclaim} examine the effect of including decomposition in fact verification pipelines, and find there exists a trade-off between accuracy and noise when using decomposition.

\subsection{Decontextualization}
Decontextualization is the technique of rewriting a sentence to stand alone, interpretable outside of the context of the passage in which it appears. Previous work \citep{choi-etal-2021-decontextualization} laid out a framework for decontextualization by editing a sentence in four different ways: (1) name completion, pronoun/NP swap, (2) discourse marker removal, (3) bridging global scope, and (4) addition. The last two edits involve adding text, including prepositional phrases or background information.

\citet{gunjal2024desiderata} examine the balance between decontextualization, making the sentence stand alone, and minimality, how little information is added. They examine this in the setting of biographies of ambiguous entities, where one name could denote multiple different entities, specifically for fact verification. 
\citet{wang-etal-2024-factcheckbench} use decontextualization as a part of an end-to-end factuality annotation solution for fact checking LLM responses.
\citet{wei2024longform} develop a method called Search-Augmented Factuality Evaluator (SAFE), for evaluating the factuality of long-form LLM responses.
\citet{lee-etal-2024-ambigdocs} build a dataset of ambiguous entities, called AmbigDocs. Although their study does not include decontextualization, this dataset is a relevant use case.

Beyond the application of fact verification, \citet{newman-etal-2023-question} use decontextualization to rewrite snippets from scientific documents to stand alone. They use three steps: (1) question generation, (2) question answering, (3) rewriting. \citet{potluri-etal-2023-concise} propose an extract-and-decontextualize approach for generating summaries of long-form answers to complex questions. \citet{kane2023gistdecontext} experiment with zero- and few-shot decontextualization using LLMs.

\begin{table*}[ht]
\small
\centering
\begin{tabular}{m{0.41\textwidth}|m{0.55\textwidth}}
    \toprule
    % \multicolumn{2}{l}{Paragraph}\\
    \multicolumn{2}{l}{Claim: He first gained recognition in the mid-1990s for his starring role in the film "Schindler's List,"} \\\multicolumn{2}{l}{directed by Steven Spielberg.}\\
    \midrule
    \textbf{Decomposition} & \textbf{Decomposition $\rightarrow$ Decontextualization} \\
    \rowcolor{gray!10}- He gained recognition in the mid-1990s. & - Liam Neeson, the actor from Northern Ireland, gained recognition in the mid-1990s. \\
    - He gained recognition for his starring role. & - Liam Neeson, the actor from Northern Ireland, gained recognition for his starring role in the film "Schindler's List." \\
    \rowcolor{gray!10}- His starring role was in the film Schindler’s List. & - Liam Neeson’s starring role was in the film Schindler’s List. \\
    - Schindler’s List is a film. & - "Schindler's List," directed by Steven Spielberg, is a film. \\
    \rowcolor{gray!10}- Steven Spielberg directed Schindler’s List. & - Steven Spielberg directed Schindler's List. \\
    - He gained recognition for his role in Schindler’s List in the mid-1990s. & - Liam Neeson gained recognition for his role in Schindler’s List in the mid-1990s. \\
    \midrule
    \textbf{Decontextualized Claim} & \textbf{Decomposition of Decontextualized Claim (Decontextualization $\rightarrow$ Decomposition)} \\
    & - Liam Neeson gained recognition in the mid-1990s. \\
    & - Liam Neeson’s recognition was for his role in Schindler's List. \\
    Liam Neeson first gained recognition in the  & - Liam Neeson had a starring role in Schindler's List. \\
    mid-1990s for his starring role in the film & - Schindler's List is a film. \\
    "Schindler's List," directed by Steven Spielberg. & - Schindler's List was directed by Steven Spielberg. \\
    & - Liam Neeson gained recognition for his role in Schindler's List in the mid-1990s. \\
    & - Liam Neeson's recognition came after Schindler's List was released. \\
    \midrule
    \multicolumn{2}{c}{\textbf{Decomposition and Decontextualization Jointly: \decompdecontext{}}} \\
    \textit{Subclaims} & \textit{Decontextualized} \\
    \rowcolor{gray!10}- He gained recognition. & - Liam Neeson gained recognition.\\
    - He gained recognition in the mid-1990s. & - Liam Neeson gained recognition in the mid-1990s. \\
    \rowcolor{gray!10}- His recognition was for a starring role. & - Liam Neeson's recognition was for a starring role. \\
    - His starring role was in the film `Schindler's List.' & - Liam Neeson's starring role was in the film 'Schindler's List.' \\
    \rowcolor{gray!10}- `Schindler's List' is a film. & - `Schindler's List' is a film directed by Steven Spielberg. \\
    - `Schindler's List' was directed by Steven Spielberg. & - `Schindler's List' was directed by Steven Spielberg. \\
    \rowcolor{gray!10}- The film `Schindler's List' contributed to his recognition. &  - The film 'Schindler's List' contributed to Liam Neeson's recognition. \\
    - The time period of the mid-1990s refers to the years around 1995. & - The time period of the mid-1990s refers to the years around 1995. \\
    \bottomrule
\end{tabular}
\caption{Examples of each method of decomposition and decontextualization. Decomposition and \decompdecontext{} subclaim sets are similar, and the decontextualized sets are similar. Decomposition is done using the \decompmethod{} method described in section \ref{sec:decomp}. We use the Molecular Facts method described in section \ref{sec:decontext} for decontextualization. We use \decompdecontext{} from section \ref{sec:dnd} for joint decomposition and decontextualization.}
\label{tab:methods-example}
\end{table*}

\section{Methods: Claim Decomposition and Decontextualization} \label{sec:d-d-methods}

In this section, we describe methods for decontextualization and decomposition which we adopt in our experiments. Sections \ref{sec:decontext} and \ref{sec:decomp} consider extenstively validated methods from prior work for decontextualization and decomposition respectively, and Section \ref{sec:dnd} proposes a prompt-based method for joint decontextualization and decomposition we call \decompdecontext{}.

\subsection{Decontextualization: Molecular Facts}\label{sec:decontext}
\citet{gunjal2024desiderata} evaluate and compare existing decontextualization methods \citep{wei2024longform} on minimality and decontextuality. They call their method ``Molecular Facts'', which they recommend above other methods of decontexualization due to its balance in minimality. This method uses a two-step prompt for decontextualizing a sentence: first using an LLM to identify ambiguities in the sentence, extracting the ambiguities in an outputted dictionary, and then prompting an LLM with these ambiguities to decontextualize the text. Because this decontextualization method was developed for human biographies, we adopt the Molecular Facts method for our experiments. We use GPT-4o mini for both the disambiguation and decontextualization step.

\subsection{Decomposition: \decompmethod{}}\label{sec:decomp}

\citet{wanner-etal-2024-closer} investigated various decomposition methods and their faithfulness, coherence, and atomicity. They introduced a method grounded in Russellian and Neo-Davidsonian theory, \decompmethod{}, and find this method to have the highest atomicity in comparison with other methods, while still remaining faithful and coherent to the original claim. Their prompt-based method uses high quality in-context decomposition examples. We use \decompmethod{} for all decompositions in our experiments. We use \texttt{gpt-3.5-turbo-instruct} for decomposition, as used in \citet{wanner-etal-2024-closer}.

\subsection{Joint Decontextualization and Decomposition: \decompdecontext{}}\label{sec:dnd}

Running decomposition and decontextualization in sequence can create problems. These two steps fundamentally interact, so considering each step independently poses limitations on the decisions made within the step. Additionally, running these in sequence doubles the number of LLM calls. 
Therefore, we develop a new method called \decompdecontext{} (Decontextualization and Decomposition) to obtain two sets of claims with just one LLM call. The first contains the decomposed subclaims of a sentence and the second set of claims contains the decontextualized form of each of the decomposed subclaims. Each claim in the subclaim set has a corresponding decontextualized claim in the decontextualized set. The prompt we developed for this method appears in Appendix \ref{app:dd-prompt}. We use GPT-4o mini for \decompdecontext{}.

\subsection{Decomposition Evaluation} \label{sec:eval}
Previous work has proposed several different ways to measure the results of a decomposition method \citep{wanner-etal-2024-closer}.
We use \decompscore{}, which measures the average number of supported (highly correlated with NLI entailment) subclaims per passage produced.
This metric indicates which method generates the most subclaims that cohere with the sentence being decomposed.

\section{Methods: Claim Verification}

We consider two decompose-then-verify methods for claim verification. First, we adopt the widely used \factscore{} from previous work. We then propose a new method that utilizes decontextualization to better isolate the specific claim to the verified.

\subsection{\factscore{}}\label{sec:factscore}
Decompose-Then-Verify metrics have become increasingly popular, including the introduction of \factscore{} \citep{min2023factscore}. \factscore{} is an LLM-based fact-checking pipeline that verifies claims decomposed from a passage against a trusted reference source (e.g., Wikipedia), and the percentage of decomposed claims supported is the \factscore{}. We use \factscore{} as a baseline, however, it is insufficient for evaluating subclaims that require more context. We use Inst-\textsc{llama} from \factscore{}, which is a \textsc{llama} 7B trained on the Super Natural Instructions dataset \citep{wang-etal-2022-super, touvron2023llamaopenefficientfoundation}, and run \factscore{} with the Inst-\textsc{llama} + retrieval + NPM setting.

\subsection{\newscore{}}\label{sec:newscore}
When a claim is augmented with context, the resulting statement contains multiple atomic facts. Passing this statement to a verification step, such as \factscore{}, can be unpredictable. Which fact is being verified? How do we ensure that the same context is not re-verified across multiple claims?

To handle these issues, we propose \newscore{}, which modifies the prompt to incorporate additional information. Specifically, the prompt is provided with the source document, the subclaim and the augmented, decontextualized claim. The method is asked to verify the specific subclaim using the relevant context against the source document. The prompt appears in Appendix \ref{app:newscore-prompt}. We use the same Inst-\textsc{llama} described in the previous section.

\section{Results}

We use the released data from \citet{min2023factscore}, which includes generated biographies from 12 language models of varying sizes. Entities for these biographies range from very rare to very frequent and span different nationalities. We treat these generations as fixed, and do not generate additional biographies or modify the existing biographies in this dataset.

\subsection{Methods of Decomposition and Decontextualization}

We use the decomposition (\decompmethod{}) and decontextualization (Molecular Facts) methods (Section \ref{sec:d-d-methods}) in sequence to evaluate the interactions of these two methods. We then evaluate our method \decompdecontext{}, using one LLM call for joint decomposition and decontextualization. We evaluate these approaches using \decompscore{} and a qualitative analysis.

The \decompscore{} results are shown in Table \ref{tab:decompscore}, with full results in Table \ref{tab:all-decompscore} in the Appendix. The \decompscore{} remains high for each of the methods, indicating a high number of facts entailed by the original sentence. Despite more information being added to decontextualized claims, we see only a small increase \decompscore{}. This suggests the added information for decontextualized claims is supported by the original claim, or a minimal enough addition that it does not change the entailment judgment.

\begin{table}[ht]
\small
\centering
\begin{tabular}{m{0.181\textwidth}|M{0.12\textwidth}|M{0.09\textwidth}}
    \toprule
    \multicolumn{1}{c|}{Method} & Avg \decompscore{} (\%) & Avg \# Subclaims \\
    \midrule
    Decomp & 95.69 & 43.48 \\ 
    Decomp $\rightarrow$ Decontext & 95.87 & 43.48 \\
    Decontext $\rightarrow$ Decomp & 96.80 & 45.48 \\
    \decompdecontext{} Subclaim & 96.14 & 36.03 \\ 	
    \decompdecontext{} Decontext & 96.58 & 36.03 \\
    \bottomrule
\end{tabular}
\caption{\decompscore{} results for each decomposition and decontextualization method. We report the two sets of claims returned by our joint approach (\decompdecontext{}) alongside both sequential approaches. Scores remain high even with the use of decontextualization. Full results can be found in table \ref{tab:all-decompscore}.}
\label{tab:decompscore}
\end{table}

Table \ref{tab:methods-example} includes an example sentence and the result of each decomposition and decontextualization method. The decomposition set and subclaims set from \decompdecontext{} contain an almost identical set of claims, indicating the new \decompdecontext{} is aligned with previously validated forms of decomposition. Similarly, the Decomposition $\rightarrow$ Decontextualization set, Decontextualization $\rightarrow$ Decomposition set, and the \decompdecontext{} decontextualized set all contain a similar set of claims, although the latter two contain more facts than the former. Because the claim is decontextualized first in Decontextualization $\rightarrow$ Decomposition, the decontextualized subclaim decomposes into more subclaims than Decomposition $\rightarrow$ Decontextualization. The decontextualized claim only replaces the pronoun and does not add any extra external information, which is expected in a simple sentence like this one.

\subsection{Fact Verification}

\begin{table*}[ht]
\small
\centering
\begin{tabular}{M{0.1\textwidth}|M{0.19\textwidth}M{0.19\textwidth}|M{0.068\textwidth}|M{0.068\textwidth}|M{0.086\textwidth}|M{0.087\textwidth}}
    \toprule
    Factuality Score & \multicolumn{2}{c|}{Decompose Method} & Avg Score (\%) & Avg \# Subclaims & Avg Score with \core{} (\%) & Avg \# Subclaims with \core{} \\
    \midrule
    \multirow{5}{*}{\factscore{}} & \multicolumn{2}{c|}{Decomp Only} & 33.00 & 43.48 & 32.27 & 23.97 \\
    & \multicolumn{2}{c|}{Decomp $\rightarrow$ Decontext} & 45.97 & 43.48 & 43.51 & 20.93 \\
    & \multicolumn{2}{c|}{Decontext $\rightarrow$ Decomp} & 44.60 & 45.48 & 40.89 & 24.08 \\
    & \multicolumn{2}{c|}{\decompdecontext{} Subclaim} & 35.92 & 36.03 & 35.43 & 22.39 \\
    & \multicolumn{2}{c|}{\decompdecontext{} Decontextualized} & 47.70 & 36.03 & 45.70 & 19.76 \\
    \midrule
    \multirow{6}{*}{\newscore{}} & Context & Verified Subclaim & & & &  \\
    \cline{2-3}
    & Original Sentence & Decomp & 41.53 & 43.48  & 41.38 & 23.97 \\
    & Decomp $\rightarrow$ Decontext & Decomp & 46.44 & 43.48 & 46.80 & 23.97 \\
    & Decontext Sentence & Decontext $\rightarrow$ Decomp & 61.51 & 45.48 & 59.18 &  24.08\\
    & \decompdecontext{} Decontextualized & \decompdecontext{} Subclaim & 51.60 & 36.03 & 51.56 & 22.39 \\
    \bottomrule
\end{tabular}
\caption{Fact verification results using different combinations of decomposition and decontextualization, and with two fact verification methods: \factscore{} and \newscore{}, a contextualized version of the former. We report the two sets of claims returned by our joint approach (\decompdecontext{}) alongside both sequential approaches. These scores are averaged across different language model splits. The full data can be found in Tables \ref{tab:all-factscore} and \ref{tab:all-newscore} in the appendix. We additionally report the deduplicated \factscore{}, \newscore{}, and average number of subclaims for each method, filtered with \core{} \citep{jiang2024corerobustfactualprecision}.}
\label{tab:fact-verification}
\end{table*}

We evaluate \factscore{} using (1) decomposition only (as in the original paper \citep{min2023factscore}), (2/3) using the decomposition and decontextualization methods in sequence (both decomposition $\rightarrow$ decontextualization and decontextualization $\rightarrow$ decomposition), and then the (4) subclaim set and (5) decontextualized set obtained from our \decompdecontext{} method. We then use our proposed fact verification method \newscore{}, which requires context for each verified subclaim. When applying \newscore{} to the decomposition-only method, we use the original sentence as context and the decomposition for the claim. For evaluating Decomposition $\rightarrow$ Decontextualization, we can use the pairs of decomposed subclaims and their corresponding decontextualized claim as context. We use the output of Decontextualization $\rightarrow$ Decomposition as the verified subclaim, and the decontextualized sentence as context. For \decompdecontext{}, we can just use the subclaim and corresponding decontextualized subclaim as context for verification.

\factscore{} and \newscore{} results are shown in Table \ref{tab:fact-verification}, with full \factscore{} results in Table \ref{tab:all-factscore} and full \newscore{} results in Table \ref{tab:all-newscore} in the Appendix. The original \factscore{} method yields an average score of 33.00\% and 43.48 average subclaims per paragraph. By design, the Decomposition $\rightarrow$ Decontextualization method has the same number of subclaims, but almost 13\% higher \factscore{}. The Decontextualization $\rightarrow$ Decomposition generates the highest number of subclaims, with more information added into the original sentence in the decontextualization step, but still achieves around the same \factscore{} as Decomposition $\rightarrow$ Decontextualization. The \decompdecontext{} subclaim set receives a similar \factscore{} as decomposition only, and the \decompdecontext{} decontextualized set has a similar \factscore{} as Decomposition $\rightarrow$ Decontextualization, but both with a smaller set of subclaims.

The \newscore{} evaluation is on average higher than the \factscore{} counterpart, while the number of subclaims is the same, by design. The \newscore{} is highest for the Decontextualization $\rightarrow$ Decomposition as verified subclaim, and decontextualized claim as context at 61.51\%. Verifying the decomposed subclaims with both the original sentence as context and the Decomposition $\rightarrow$ Decontextualization as context achieves similar scores, within five percent of one another. Evaluating the \decompdecontext{} method subclaim and decontextualization sets with \newscore{} results in a score in between the others, at 51.60\%.

We show that factuality scores change when using \factscore{} evaluated decontextualized subclaims instead of decomposed subclaims, however using only these decontextualized claims is insufficient. \newscore{} addresses the ambiguity problems of using decomposed subclaims, and the redundancy and loss of atomicity problems of using decontextualized subclaims. The \factscore{} and \newscore{} evaluations in table \ref{tab:fact-verification} demonstrate that these factuality scores are changing. In the following section, we aim to understand \textit{what} causes these changes.

\section{Analysis}

\subsection{Verification Changes: Quantitative Evaluation}

Based on the change in factuality scores, we aim to understand: what makes a claim verifiable and what causes the changes of these factuality scores for the same passage? We examine the \newscore{} evaluation, and \decompdecontext{} \factscore{} evaluation of the subclaim set and decontextualization set. We specifically study these results, because we can align at a subclaim level and examine the different subclaims and corresponding decontextualized claims evaluated by \factscore{}, and the pair evaluated by \newscore{}. Additionally, the decomposition only and the \decompdecontext{} subclaim set achieve a similar factuality score, and the Decomposition $\rightarrow$ Decontextualization achieves a similar factuality score as \decompdecontext{} decontextualized set, and therefore we expect these sets of subclaims are similar.

We first examine the percentage of claims whose support judgment changed. We find that between \decompdecontext{} subclaim and \decompdecontext{} decontextualization \factscore{} evaluated sets, there is a 19.11\% change in judgment, 16.25\% change from false to true when decontextualizing. 48.52\% of these contain a pronoun replacement\footnote{Subclaims which contain one or more common pronouns ("she", "her", "hers", "herself", "he", "him", "his", "himself", "they", "them", "theirs", and "themself"), which was not found in the decontextualized claim are indicated as having pronoun replacement.}, suggesting entity disambiguation is a big reason for judgments to switch from false to true. The additional information helps verification. \decompdecontext{} subclaim and \decompdecontext{} decontextualization \factscore{} evaluation changes from true to false only 3.26\% of the time, 11.82\% of these containing pronoun replacement. Incorrectly verified claims were less of an issue.

Between the \decompdecontext{} subclaim \factscore{} evaluations and \newscore{} evaluations, 16.97\% of subclaim judgments change, with 16.50\% changing from false to true, and 0.48\% changing from true to false. The addition of context rarely changes judgments to false, and instead only adds additional background or pronoun replacement that makes the claim easier to verify as correct.

\subsection{Verification Changes: Qualitative Evaluation}

\begin{table*}[ht]
\small
\centering
\begin{tabular}{cm{0.59\textwidth}cc}
    \toprule
    \textbf{Example} & \multicolumn{1}{c}{\textbf{Subclaims}} & \textbf{\factscore{}} & \textbf{\newscore{}} \\
    \midrule
    \multirow{2}{*}{1} & \textbf{Subclaim}: Prince Daniel is a member of the Swedish royal family.  & False & \multirow{2}{*}{True} \\
    & \textbf{Decontextualized}: Prince Daniel, who is a sibling of Prince Carl Philip, is a member of the Swedish royal family. & True &  \\
    \midrule
    \multirow{2}{*}{2} & \textbf{Subclaim}: She has appeared in several television series. & False & \multirow{2}{*}{True} \\
     & \textbf{Decontextualized}: Susan Sarandon has appeared in several television series. & True & \\
    \midrule
    \multirow{2}{*}{3} & \textbf{Subclaim}: The sitcom featured Matthew Perry as a lead actor. & True & \multirow{2}{*}{False} \\
    & \textbf{Decontextualized}: `The Matt Payne Show' featured Matthew Perry as a lead actor. & False &\\
    \midrule
    \multirow{2}{*}{4} & \textbf{Subclaim}: She wrote for the school's newspaper. & True & \multirow{2}{*}{False} \\
     & \textbf{Decontextualized}: Nikole Hannah-Jones wrote for the newspaper of Wesleyan University. & False &\\
    \midrule
    \multirow{2}{*}{5} & \textbf{Subclaim}: He began his wrestling career. & True & \multirow{2}{*}{True}\\
    & \textbf{Decontextualized}: Fuerza Guerrera, also known as Juan Conrado Aguilar Jáuregui, began his wrestling career. & False & \\
    \bottomrule
\end{tabular}
\caption{Examples of subclaims and their decontextualized claim as generated by \decompdecontext{}, and their \factscore{} evaluations, and the \newscore{} evalutation for the pair of claims. Context helps evaluate the factuality of claims, and \newscore{} can handle cases where incorrect context is added to the text. Examples 1 and 3 show decontextualization with added information, and examples 2, 4, and 5 demonstrate pronoun replacement.}
\label{tab:judgement-analysis}
\end{table*}

Context is important for fact verification, shown in Table \ref{tab:judgement-analysis}. Here we look at \decompdecontext{} subclaim and decontextualized subclaim pairs each evaluated separately with \factscore{}, and corresponding \newscore{} judgments. We present examples where factuality judgments disagree, demonstrate where addition of context helps disambiguate the claim, and show how \newscore{} can handle these cases.

Example 1 in Table \ref{tab:judgement-analysis} shows an example where judgment changes from false to true, without pronoun replacement. ``Prince Daniel'' is a name which could refer to the current member of the Swedish royal family (born 1973), the prince of Galicia (1201–1264), the Russian prince from 1261–1303, or Prince Daniel of Saxony (born 1975). The added information to the verified subclaim disambiguates which ``Prince Daniel'' is the subject of the paragraph. Example 2 shows an example where \factscore{} validation changes from false to true \textit{with pronoun replacement}. The pronoun ``She'' is replaced with ``Susan Sarandon'' disambiguating the subject of the sentence. The \newscore{} for the first two examples is also correctly judged as true with context.

Examples 3 and 4 in Table \ref{tab:judgement-analysis} had a \factscore{} judgment changed from true to false. The third example, with no pronoun replacement, disambiguates which sitcom is referenced, changing if the subclaim is verified\footnote{Matthew Perry was not the lead actor in the `The Matt Payne Show'}. The fourth example replaces ``She'' with ``Nikole Hannah-Jones''. Both of these claims are correctly judged as false using \newscore{}.

There is a lot of movement between the \factscore{} judgments of subclaims and the \factscore{} judgments of decontextualized claims. The additional context can help, as in the examples shown, however, there exist examples where decontextualization is not sufficient. In some cases added contextual information is wrong, and masks the correctness of the subclaim being verified. Example 5 in Table \ref{tab:judgement-analysis} considers an example about Mexican wrestler Fuerza Guerrera where the atomic fact is true, but the information added to the decontextualized claim is false.\footnote{Juan Conrado Aguilar Jáuregui is the name of a different Mexican wrestler, and is not another name for Fuerza Guerrera.} The \factscore{} evaluation of the subclaim is true, but false for the decontextualized claim, because ``Conrado Aguilar Jáuregui'' refers to a different Mexican wrestler. Despite this incorrect addition to the decontextualized subclaim, \newscore{} evaluates the subclaim as true. \newscore{} verifies the subclaim, while still including the context. This ensures that the context helps to verify claims, but does not overshadow the subclaim being verified. \newscore{} can handle these nuances lending it to be a more robust method for fact verification.

\subsection{\core{} for Subclaim Deduplication}

\citet{jiang2024corerobustfactualprecision} presents \core{}, a filtering method for sets of decomposed subclaims. \core{} formats subclaim subselection as a constrained optimization problem to eliminate duplicated facts, which we apply at a generation level. Deduplication results can be found in Table \ref{tab:fact-verification}. We use this filtering as an analysis of the different methods of decontextualization, decomposition, and verification, however, it can (and should) be used to filter sets subclaims for verification to ensure scores are not inflated with many duplicated facts.

We find the average set of filtered subclaims for all decomposition and decontextualization methods to be between 19 and 25. The size of the filtered subclaims have less variance between them because \core{} filters to a \textit{core} set of subclaims. The \decompdecontext{} decontextualized set had more subclaims removed compared to the \decompdecontext{} subclaim set, indicating a more redundant decontextualized set. As shown in Figure \ref{fig:motivation}, the decontextualized form of subclaims are less atomic and include information from other subclaims. The redundant subclaims are removed in the deduplication process.

Filtering the subclaim \decompdecontext{} set results in the least number of facts, and decontextualize-then-decompose have the most remaining subclaims, due to added context having slightly different wording, and not fully being filtered out. For example, in the set of decontextualized-then-decomposed subclaims from Vicuna-7B on Dr. Dre, multiple claims stating something equivalent to ``Dr. Dre was a rapper'' were kept in the filter process, despite being semantically equivalent. This is an example of the repeated essential context added to each decontextualized sentence in the paragraph before decomposing as shown in the bottom right of Figure \ref{fig:motivation}. The \decompdecontext{} sets contained fewer subclaims, and thus had fewer subclaims removed in the filtering process. 
\factscore{} changes are within two percent for all deduplicated subclaim sets, except for decontextualize-then-decompose which drops almost four percent. As shown in Figure \ref{fig:motivation}, this method is at risk of containing duplicated essential context across sentences.

\section{Conclusion}

In this work, we consider the interactions of decomposition and decontextualization in a fact verification system. We introduce \decompdecontext{}, a prompt-based method for extracting subclaims and corresponding decontextualized forms. 
Using pairs of subclaims and decontextualized claims, we propose a new decompose-then-verify method, \newscore{}, which validates claims with a given context. We demonstrate cases where verification judgment differs, and show \newscore{} is able to handle context better than previous measures of factuality.

\section{Limitations}

Although we demonstrate the robustness of \newscore{} in handling ambiguities, which is especially important in verification of generations with many entities, we only consider its use in fact verification of generated biographies. Our work does not show the use of \newscore{} 

\newscore{} is not intended to handle debatable claims, such as opinions. Previous work has examined factuality scores of only verifiable claims \citep{song-etal-2024-veriscore}, which could be integrated into \newscore{}, such that \newscore{} only evaluates these verifiable claims.

The reference documents are static in this study, but using decontextualized subclaims for reference document retrieval is a possible line of future work, that would allow this method to work on arbitrary generations, instead of generations on a predetermined topic.

A important limitation of our evaluation is the underlying dataset, a set of biographies for which the correct source (Wikipedia) documents are available for verification. In a production system, we may find source document selection errors due to the information retrieval process. We hypothesize that these will increase verification confusions, and the differences between our approaches and previous work will grow. Additionally, while biographies center on a single individual, other generation types, e.g. news updates, may include multiple entities and increase verification confusion.

This work is only done in the language of English, although we expect these results would hold in other languages.

\section{Ethics Statement}
LLMs are prone to hallucination, which can result in the spread of misinformation. Mitigating these hallucinations is important and still actively being researched. The decomposition and decontextualization methods in this work are at risk of injecting hallucinated information not in the original claim. Evaluation of these untrue generations are necessary to ensure the reliability of them, and care should be taken when trusting these generations.

\bibliographystyle{acl_natbib}
\bibliography{main}

\clearpage
\newpage

\appendix

\section{Prompts}
\subsection{Decomposition Prompt}

We use the \decompmethod{} prompt from \citet{wanner-etal-2024-closer}. Their prompt uses dynamically retrieved in-context decompositions.

\subsection{Decontextualization Prompt} 

We use the Molecular Facts prompt from \citet{gunjal2024desiderata}. They use a two step prompting methods, with the first prompt identifying ambiguities in a claim, and the second prompt using the identified ambiguities to decontextualize the claim.

\subsection{\decompdecontext{} Prompt}\label{app:dd-prompt}
The \decompdecontext{} prompt for extracting pairs of subclaims and their decontextualized form can be found in Tables \ref{tab:first-d+d-prompt}-\ref{tab:last-d+d-prompt}. We use the ambiguity criteria outlined in the Molecular Facts prompt and decomposed in-context examples from \decompmethod{}.

\begin{table*}[h!]
\centering
\begin{tabular}{p{12cm}}
\hline
\scriptsize
\begin{verbatim}
Ambiguity Criteria: Ambiguity manifests in diverse forms, including:
- Similar names denoting distinct entities.
- Varied interpretations stemming from insufficient information.
- Multiple understandings arising from vague or unclear information.

Instructions:
- You are given a paragraph, and one sentence from the paragraph to decompose and decontextualize.
- First decompose the sentence into subclaims. Only use information from the sentence, and do not 
add any external information.
- Then using those subclaims, write a decontextualized version of each subclaim.
- In the decontextualized version, include all necessary information to disambiguate any entities 
or events in the subclaim using the ambiguity criteria above.
- In the decontextualized version, only use information from the paragraph. Do not add any external
information.
- Provide an explanation of what ambiguities need to be resolved

Format your response as a combination of decomposition and a dictionary with pairs of context and 
subclaims:
##PARAGRAPH##: <paragraph>
##SENTENCE##: <sentence>
##SUBCLAIMS##:
<list-of-subclaims>
##EXPLANATION##:
<explanations>
##CONTEXT-SUBCLAIM PAIRS##:
[
    {"subclaim": <subclaim1>, "decontextualized": <context1>},
    {"subclaim": <subclaim2>, "decontextualized": <context2>},
    ...
]

Example 1:
##PARAGRAPH##: Michael Collins (born October 31, 1930) is a retired American astronaut and test 
pilot who was the Command Module Pilot for the Apollo 11 mission in 1969. He orbited the Moon 
in the command module Columbia while Neil Armstrong and Buzz Aldrin made their historic landing. 
Born in Rome, Italy, Collins graduated from the U.S. Military Academy in 1952, joining a family 
tradition of military service, and went on to become a test pilot in the U.S. Air Force. Selected
as an astronaut in 1963, he flew two space missions, Gemini 10 in 1966 and Apollo 11 in 1969, 
making him one of only 24 people to travel to the Moon. Collins was an accomplished astronaut, 
becoming the fourth person to conduct a spacewalk and the first to perform multiple spacewalks. 
After leaving NASA in 1970, he served as Assistant Secretary of State for Public Affairs, later 
directing the National Air and Space Museum. He also held senior roles at the Smithsonian and in 
private aerospace, eventually founding his own consulting firm. Collins and his Apollo 11 crewmates 
received the Presidential Medal of Freedom in 1969 and the Congressional Gold Medal in 2011.
##SENTENCE##: Michael Collins (born October 31, 1930) is a retired 
American astronaut and test pilot who was the Command Module Pilot 
for the Apollo 11 mission in 1969.
##SUBCLAIMS##:
- Michael Collins was born in October.
- Michael Collins was born on the 31st day of a month.
- Michael Collins was born in 1930.
- Michael Collins is retired.
- Michael Collins is American.
- Michael Collins was an astronaut.
- Michael Collins was a test pilot.
- Michael Collins participated in the Apollo 11 mission.
- Michael Collins's participation in the Apollo 11 mission occurred in 1969.
- The Apollo 11 mission was active in 1969.
- The day of Michael Collins's birth occurred before his year of participation 
in the Apollo 11 mission.
- The Apollo 11 mission had a Command Module Pilot.
- Michael Collins's role in the Apollo 11 mission was as the Command Module Pilot.
##EXPLANATION##:
"Michael Collins" needs to be disambiguated as the astronaut associated with the 
Apollo 11 mission to distinguish him from other potential individuals with 
similar names.
\end{verbatim}\\
\hline
\end{tabular}
\caption{1/3 of the \decompdecontext{} method for extracting subclaims and corresponding decontextualized subclaims. More details can be found in appendix section \ref{app:dd-prompt}.}
\label{tab:first-d+d-prompt}
\end{table*}

\begin{table*}[h!]
\centering
\begin{tabular}{p{12cm}}
\hline
\scriptsize
\begin{verbatim}
##CONTEXT-SUBCLAIM PAIRS##:
[
    {"subclaim": "Michael Collins was born in October.", 
     "decontextualized": "Michael Collins, the retired American astronaut and test pilot, was born 
     in October."},
    {"subclaim": "Michael Collins was born on the 31st day of a month.", 
     "decontextualized": "Michael Collins, the retired American astronaut and test pilot, was born 
     on the 31st day of a month."},
    {"subclaim": "Michael Collins was born in 1930.", 
     "decontextualized": "Michael Collins, the retired American astronaut and test pilot, was born 
     in 1930."},
    {"subclaim": "Michael Collins is retired.", 
     "decontextualized": "Michael Collins, the retired American astronaut and test pilot, is 
     retired."},
    {"subclaim": "Michael Collins is American.", 
     "decontextualized": "Michael Collins, the American astronaut, is American."},
    {"subclaim": "Michael Collins was an astronaut.", 
     "decontextualized": "Michael Collins, the retired American astronaut and Command Module Pilot 
     for the Apollo 11 mission, was an astronaut."},
    {"subclaim": "Michael Collins was a test pilot.", 
     "decontextualized": "Michael Collins, the retired American astronaut and test pilot, was the 
     Command Module Pilot for the Apollo 11 mission in 1969."},
    {"subclaim": "Michael Collins participated in the Apollo 11 mission.", 
     "decontextualized": "Michael Collins, the retired American astronaut and test pilot, 
     participated in the Apollo 11 mission."},
    {"subclaim": "Michael Collins's participation in the Apollo 11 mission occurred in 1969.", 
     "decontextualized": "Michael Collins's participation in the Apollo 11 mission as the Command 
     Module Pilot occurred in 1969."},
    {"subclaim": "The Apollo 11 mission was active in 1969.", 
     "decontextualized": "The Apollo 11 mission, which involved human spaceflight to the Moon, was 
     active in 1969."},
    {"subclaim": "The day of Michael Collins's birth occurred before his year of participation in 
    the Apollo 11 mission.", 
     "decontextualized": "The day of Michael Collins's birth on October 31, 1930, occurred before 
     his year of participation in the Apollo 11 mission."},
    {"subclaim": "The Apollo 11 mission had a Command Module Pilot.", 
     "decontextualized": "The Apollo 11 mission had Michael Collins as its Command Module Pilot."},
    {"subclaim": "Michael Collins's role in the Apollo 11 mission was as the Command Module 
    Pilot.", 
     "decontextualized": "Michael Collins's role in the Apollo 11 mission was as the Command Module 
     Pilot."}
]

Example 2:
##PARAGRAPH##: Stephen Miller (born August 23, 1985) is an American political advisor who served 
as a senior advisor for policy and director of speechwriting to President Donald Trump. Miller has 
been described as the architect of Trump's controversial immigration policies, and has previously 
worked for Alabama Senator Jeff Sessions on immigration issues. Miller was instrumental in shaping 
several of Trump's key policies, including the travel ban, a reduction in refugee admissions, and 
family separations at the border. He began his career in communications roles for conservative 
legislators, including Senators Jeff Sessions, Michele Bachmann, and John Shadegg. As Trump's 
speechwriter, Miller helped draft the inaugural address and served as a trusted advisor from the 
early days of the administration. He also played a significant role in the resignation of Secretary 
of Homeland Security Kirstjen Nielsen, whom he deemed insufficiently strict on immigration. As a 
White House spokesperson, Miller made several unsubstantiated claims about election fraud and 
promoted content from white nationalist sources, leading to his inclusion on the Southern Poverty 
Law Center's list of extremists.
\end{verbatim}\\
\hline
\end{tabular}
\caption{2/3 of the \decompdecontext{} method for extracting subclaims and corresponding decontextualized subclaims. More details can be found in appendix section \ref{app:dd-prompt}.}
\end{table*}

\begin{table*}[h!]
\centering
\begin{tabular}{p{12cm}}
\hline
\scriptsize
\begin{verbatim}
##SENTENCE##: Miller has been described as the architect of Trump's controversial immigration 
policies, and has previously worked for Alabama Senator Jeff Sessions on immigration issues.
##SUBCLAIMS##:
- Miller has been described.
- Miller has been described as an architect.
- Miller has been described as an architect of Trump's controversial 
immigration policies.
- Trump has immigration policies.
- Trump's immigration policies are controversial.
- Miller worked for Jeff Sessions.
- Jeff Sessions is a Senator.
- Jeff Sessions represents Alabama.
- Miller worked on immigration issues.
- Miller's work for Jeff Sessions involved immigration issues.
##EXPLANATION##:
"Miller" needs to be disabiguated as Stephen Miller, a political advisor for Donald Trump, to avoid 
confusion with other individuals with the same name. Clarify that "Trump's immigration policies" 
refers specifically to policies developed during Donald Trump's presidency, as "Trump" alone may be 
ambiguous in a different context.
##CONTEXT-SUBCLAIM PAIRS##:
[
    {"subclaim": "Miller has been described.", 
     "decontextualized": "Miller, the architect of Trump's controversial immigration policies, has 
     been described."},
    {"subclaim": "Miller has been described as an architect.", 
     "decontextualized": "Miller, who has been described as the architect of Trump's controversial 
     immigration policies, has been described as an architect."},
    {"subclaim": "Miller has been described as an architect of Trump's controversial immigration 
    policies.", 
     "decontextualized": "Stephen Miller has been described as an architect of Trump's 
     controversial immigration policies."},
    {"subclaim": "Trump has immigration policies.", 
     "decontextualized": "Donald Trump has immigration policies."},
    {"subclaim": "Trump's immigration policies are controversial.", 
     "decontextualized": "Donald Trump's immigration policies are controversial."},
    {"subclaim": "Miller worked for Jeff Sessions.", 
     "decontextualized": "Miller, the architect of Trump's controversial immigration policies, 
     worked for Jeff Sessions."},
    {"subclaim": "Jeff Sessions is a Senator.", 
     "decontextualized": "Jeff Sessions is a Senator from Alabama."},
    {"subclaim": "Jeff Sessions represents Alabama.", 
     "decontextualized": "Jeff Sessions represents the state of Alabama."},
    {"subclaim": "Miller worked on immigration issues.", 
     "decontextualized": "Miller, the architect of Trump's controversial immigration policies, 
     worked on immigration issues."},
    {"subclaim": "Miller's work for Jeff Sessions involved immigration issues.", 
     "decontextualized": "Stephen Miller's work for Jeff Sessions involved immigration issues."},
]

Your task:
##PARAGRAPH##: [paragraph]
##SENTENCE##: [sentence]
##SUBCLAIMS##:
\end{verbatim}\\
\hline
\end{tabular}
\caption{3/3 of the \decompdecontext{} method for extracting subclaims and corresponding decontextualized subclaims. More details can be found in appendix section \ref{app:dd-prompt}.}
\label{tab:last-d+d-prompt}
\end{table*}

\subsection{\factscore{} Prompt}

We use the \factscore{} prompt from \citet{min2023factscore} shown below. This prompt provides a reference document and asks if an atomic fact is true or false.

{\footnotesize
\begin{verbatim}
Answer the question about [TOPIC] based on the given 
context.

Title: [REFERENCE DOC SECTION TITLE]
Text: [REFERENCE DOC CONTENT]

Input: [ATOM] True or False?
Output:
\end{verbatim}
}

\subsection{\newscore{} Prompt}\label{app:newscore-prompt}
The following is the \newscore{} prompt used for factuality verification of claims given the decontextualized form of the claim. This prompt is adapted from \factscore{}, but includes the decontextualized claim as context for the atomic claim.

{\footnotesize
\begin{verbatim}
Answer the question about [TOPIC] based on the given 
reference document and context.

Reference Document:
[REFERENCE DOC]

Given the following context: "[DECONTEXT CLAIM]"
Input: Is "[ATOM]" True or False?
Output:
\end{verbatim}
}

\subsection{Compute}
Decomposition and decontextualization experiments were run on a GPU cluster with Quadro RTX 6000. We estimate experiments took around 400 GPU-hours.

\section{Full Results}
\subsection{\factscore{} Results}
The full \factscore{} results for each language model split are in Table \ref{tab:all-factscore}.

\begin{table*}[ht]
\small
\centering
\begin{tabular}{M{0.19\textwidth}|M{0.15\textwidth}|M{0.14\textwidth}|M{0.13\textwidth}}
    \toprule
    Decompose Method & LM Split & \factscore{} (\%) & \# Subclaims \\
    \midrule
    \multirow{12}{*}{Decomp Only} & Alpaca 7B & 35.0 & 22.2 \\
    & Alpaca 13B & 38.9 & 22.0 \\
    & Alpaca 65B & 44.0 & 22.2 \\
    & ChatGPT & 48.2 & 44.2 \\
    & Dolly 12B & 16.5 & 33.0 \\
    & GPT-4 & 51.1 & 77.7 \\
    & InstructGPT & 40.1 & 36.3 \\
    & MPT-Chat 7B & 24.8 & 49.0 \\
    & Oasst-pythia 12B & 20.1 & 57.7 \\
    & StableLM 7B & 13.8 & 40.4 \\
    & Vicuna 7B & 32.4 & 59.8 \\
    & Vicuna 13B & 31.1 & 57.3 \\
    \midrule
    \multirow{12}{*}{Decomp $\rightarrow$ Decontext} & Alpaca 7B & 49.8 & 22.2 \\
    & Alpaca 13B & 52.8 & 22.0 \\
    & Alpaca 65B & 60.6 & 22.2 \\
    & ChatGPT & 66.6 & 44.2 \\
    & Dolly 12B & 22.3 & 33.0 \\
    & GPT-4 & 71.8 & 77.7 \\
    & InstructGPT & 57.0 & 36.3 \\
    & MPT-Chat 7B & 33.7 & 49.0 \\
    & Oasst-pythia 12B & 29.4 & 57.7 \\
    & StableLM 7B & 17.8 & 40.4 \\
    & Vicuna 7B & 46.5 & 59.8 \\
    & Vicuna 13B & 43.4 & 57.3 \\  
    \midrule
    \multirow{12}{*}{Decontext $\rightarrow$ Decomp} & Alpaca 7B & 48.3 & 24.2 \\
    & Alpaca 13B & 51.2 & 23.7 \\
    & Alpaca 65B & 57.2 & 24.1 \\
    & ChatGPT & 61.6 & 45.5 \\
    & Dolly 12B & 24.1 & 34.7 \\
    & GPT-4 & 64.6 & 79.9 \\
    & InstructGPT & 53.5 & 38.4 \\
    & MPT-Chat 7B & 33.9 & 50.8 \\
    & Oasst-pythia 12B & 31.0 & 59.1 \\
    & StableLM 7B & 21.4 & 42.5 \\
    & Vicuna 7B & 45.1 & 61.2 \\
    & Vicuna 13B & 43.3 & 61.7 \\
    \midrule
    \multirow{12}{*}{\decompdecontext{} Subclaim} & Alpaca 7B & 36.8 & 20.7 \\
    & Alpaca 13B & 41.3 & 20.0 \\
    & Alpaca 65B & 46.5 & 20.2 \\
    & ChatGPT & 53.6 & 37.9 \\
    & Dolly 12B & 18.1 & 30.1 \\
    & GPT-4 & 56.3 & 61.8 \\
    & InstructGPT & 43.4 & 31.7 \\
    & MPT-Chat 7B & 26.5 & 40.6 \\
    & Oasst-pythia 12B & 21.5 & 43.1 \\
    & StableLM 7B & 15.1 & 33.5 \\
    & Vicuna 7B & 35.9 & 47.5 \\
    & Vicuna 13B & 36.0 & 45.2 \\
    \midrule
    \multirow{12}{*}{\decompdecontext{} Decontextualized} & Alpaca 7B & 52.0 & 20.7 \\
    & Alpaca 13B & 54.4 & 20.0 \\
    & Alpaca 65B & 62.3 & 20.2 \\
    & ChatGPT & 69.2 & 37.9 \\
    & Dolly 12B & 23.8 & 30.1 \\
    & GPT-4 & 73.7 & 61.8 \\
    & InstructGPT & 59.8 & 31.7 \\
    & MPT-Chat 7B & 34.9 & 40.6 \\
    & Oasst-pythia 12B & 29.7 & 43.1 \\
    & StableLM 7B & 18.0 & 33.5 \\
    & Vicuna 7B & 48.3 & 47.5 \\
    & Vicuna 13B & 46.3 & 45.2 \\
    \bottomrule
\end{tabular}
\caption{The \factscore{} for different language model splits, the results of which are aggregated in Table \ref{tab:fact-verification}. The factuality ranking of these language models stays consistent, however the scores differ.}
\label{tab:all-factscore}
\end{table*}

\subsection{\newscore{} Results}
The full \factscore{} results for each language model split are in Table \ref{tab:all-newscore}.

\begin{table*}[ht]
\small
\centering
\begin{tabular}{M{0.19\textwidth}|M{0.15\textwidth}|M{0.14\textwidth}|M{0.13\textwidth}}
    \toprule
    Decompose Method & LM Split & \newscore{} (\%) & \# Subclaims \\
    \midrule
    \multirow{12}{0.19\textwidth}{\textbf{Context:} Original Sentence \\\textbf{Verified Subclaim:} Decomp} & Alpaca 7B & 43.6 & 22.2 \\
    & Alpaca 13B & 46.8 & 22.0 \\
    & Alpaca 65B & 51.0 & 22.2 \\
    & ChatGPT & 53.8 & 44.2 \\
    & Dolly 12B & 25.6 & 33.0 \\
    & GPT-4 & 57.8 & 77.7 \\
    & InstructGPT & 47.8 & 36.3 \\
    & MPT-Chat 7B & 36.8 & 49.0 \\
    & Oasst-pythia 12B & 31.5 & 57.7 \\
    & StableLM 7B & 22.0 & 40.4 \\
    & Vicuna 7B & 41.7 & 59.8 \\
    & Vicuna 13B & 40.0 & 57.3 \\
    \midrule
    \multirow{12}{0.19\textwidth}{\textbf{Context:} Decomp $\rightarrow$ Decontext \\\textbf{Verified Subclaim:} Decomp} & Alpaca 7B & 49.9 & 22.2 \\
    & Alpaca 13B & 52.7 & 22.0 \\
    & Alpaca 65B & 57.9 & 22.2 \\
    & ChatGPT & 58.9 & 44.2 \\
    & Dolly 12B & 30.6 & 33.0 \\
    & GPT-4 & 62.6 & 77.7 \\
    & InstructGPT & 53.9 & 36.3 \\
    & MPT-Chat 7B & 40.7 & 49.0 \\
    & Oasst-pythia 12B & 33.8 & 57.7 \\
    & StableLM 7B & 25.3 & 40.4 \\
    & Vicuna 7B & 46.0 & 59.8 \\
    & Vicuna 13B & 45.0 & 57.3 \\
    \midrule
    \multirow{12}{0.19\textwidth}{\textbf{Context:} Decontext Sentence\\\textbf{Verified Subclaim:} Decontext $\rightarrow$ Decomp} & Alpaca 7B & 68.2 & 24.2 \\
    & Alpaca 13B & 68.8 & 23.7 \\
    & Alpaca 65B & 73.1 & 24.1 \\
    & ChatGPT & 69.6 & 45.5 \\
    & Dolly 12B & 46.5 & 34.7 \\
    & GPT-4 & 73.5 & 79.9 \\
    & InstructGPT & 70.2 & 38.4 \\
    & MPT-Chat 7B & 55.8 & 50.8 \\
    & Oasst-pythia 12B & 54.8 & 59.1 \\
    & StableLM 7B & 38.9 & 42.5 \\
    & Vicuna 7B & 62.4 & 61.2 \\
    & Vicuna 13B & 56.3 & 61.7 \\
    \midrule
    \multirow{12}{0.19\textwidth}{\textbf{Context:} \decompdecontext{} Decontextualized\\\textbf{Verified Subclaim:} \decompdecontext{} Subclaim} & Alpaca 7B & 53.9 & 20.7 \\
    & Alpaca 13B & 57.0 & 20.0 \\
    & Alpaca 65B & 61.5 & 20.2 \\
    & ChatGPT & 65.4 & 37.9 \\
    & Dolly 12B & 37.2 & 30.1 \\
    & GPT-4 & 69.4 & 61.8 \\
    & InstructGPT & 58.6 & 31.7 \\
    & MPT-Chat 7B & 44.9 & 40.6 \\
    & Oasst-pythia 12B & 38.3 & 43.1 \\
    & StableLM 7B & 29.4 & 33.5 \\
    & Vicuna 7B & 51.7 & 47.5 \\
    & Vicuna 13B & 51.9 & 45.2 \\
    \bottomrule
\end{tabular}
\caption{The \newscore{} for different language model splits, the results of which are aggregated in Table \ref{tab:fact-verification}. The factuality ranking of these language models stays consistent, however the scores differ.}
\label{tab:all-newscore}
\end{table*}

\subsection{\decompscore{} Results}
The full \factscore{} results for each language model split are in Table \ref{tab:all-decompscore}.

\begin{table*}[ht]
\small
\centering
\begin{tabular}{M{0.19\textwidth}|M{0.15\textwidth}|M{0.14\textwidth}|M{0.13\textwidth}}
    \toprule
    Decompose Method & LM Split & \decompscore{} (\%) & \# Subclaims \\
    \midrule
    \multirow{12}{*}{Decomp Only} & Alpaca 7B & 98.7 & 22.2 \\
    & Alpaca 13B & 98.6 & 22.0 \\
    & Alpaca 65B & 98.6 & 22.2 \\
    & ChatGPT & 93.0 & 44.2 \\
    & Dolly 12B & 97.4 & 33.0 \\
    & GPT-4 & 96.2 & 77.7 \\
    & InstructGPT & 98.1 & 36.3 \\
    & MPT-Chat 7B & 96.5 & 49.0 \\
    & Oasst-pythia 12B & 98.3 & 57.7 \\
    & StableLM 7B & 89.2 & 40.4 \\
    & Vicuna 7B & 94.8 & 59.8 \\
    & Vicuna 13B & 88.9 & 57.3 \\
    \midrule
    \multirow{12}{*}{Decomp $\rightarrow$ Decontext} & Alpaca 7B & 99.1 & 22.2 \\
    & Alpaca 13B & 99.0 & 22.0 \\
    & Alpaca 65B & 98.9 & 22.2 \\
    & ChatGPT & 92.9 & 44.2 \\
    & Dolly 12B & 97.8 & 33.0 \\
    & GPT-4 & 96.3 & 77.7 \\
    & InstructGPT & 98.6 & 36.3 \\
    & MPT-Chat 7B & 96.3 & 49.0 \\
    & Oasst-pythia 12B & 98.5 & 57.7 \\
    & StableLM 7B & 88.1 & 40.4 \\
    & Vicuna 7B & 95.3 & 59.8 \\
    & Vicuna 13B & 89.7 & 57.3 \\
    \midrule
    \multirow{12}{*}{Decontext $\rightarrow$ Decomp} & Alpaca 7B & 99.0 & 24.2 \\
    & Alpaca 13B & 99.1 & 23.7 \\
    & Alpaca 65B & 99.1 & 24.1 \\
    & ChatGPT & 95.1 & 45.5 \\
    & Dolly 12B & 98.1 & 34.7 \\
    & GPT-4 & 96.8 & 79.9 \\
    & InstructGPT & 98.5 & 38.4 \\
    & MPT-Chat 7B & 97.0 & 50.8 \\
    & Oasst-pythia 12B & 98.5 & 59.1 \\
    & StableLM 7B & 92.4 & 42.5 \\
    & Vicuna 7B & 96.4 & 61.2 \\
    & Vicuna 13B & 91.6 & 61.7 \\
    \midrule
    \multirow{12}{*}{\decompdecontext{} Subclaim} & Alpaca 7B & 98.9 & 20.7 \\
    & Alpaca 13B & 99.0 & 20.0 \\
    & Alpaca 65B & 99.1 & 20.2 \\
    & ChatGPT & 92.7 & 37.9 \\
    & Dolly 12B & 98.1 & 30.1 \\
    & GPT-4 & 96.0 & 61.8 \\
    & InstructGPT & 98.9 & 31.7 \\
    & MPT-Chat 7B & 97.1 & 40.6 \\
    & Oasst-pythia 12B & 99.1 & 43.1 \\
    & StableLM 7B & 89.8 & 33.5 \\
    & Vicuna 7B & 95.6 & 47.5 \\
    & Vicuna 13B & 89.4 & 45.2 \\
    \midrule
    \multirow{12}{*}{\decompdecontext{} Decontextualized} & Alpaca 7B & 99.4 & 20.7 \\
    & Alpaca 13B & 99.3 & 20.0 \\
    & Alpaca 65B & 99.4 & 20.2 \\
    & ChatGPT & 94.0 & 37.9 \\
    & Dolly 12B & 98.6 & 30.1 \\
    & GPT-4 & 96.4 & 61.8 \\
    & InstructGPT & 99.4 & 31.7 \\
    & MPT-Chat 7B & 96.9 & 40.6 \\
    & Oasst-pythia 12B & 99.1 & 43.1 \\
    & StableLM 7B & 88.4 & 33.5 \\
    & Vicuna 7B & 96.7 & 47.5 \\
    & Vicuna 13B & 91.3 & 45.2 \\
    \bottomrule
\end{tabular}
\caption{The \decompscore{} for different language model splits, the results of which are aggregated in Table \ref{tab:decompscore}. The \decompscore{} remains high despite additional new information for decontextualized subclaims.}
\label{tab:all-decompscore}
\end{table*}

\end{document}